\pdfoutput=1

\PassOptionsToPackage{dvipsnames}{xcolor}

\documentclass[11pt]{article}

\usepackage[preprint]{acl}

\usepackage{times}
\usepackage{latexsym}

\usepackage[T1]{fontenc}
\usepackage{graphicx}
\usepackage{booktabs}

\usepackage{amsmath,amsfonts,amssymb}
\usepackage{bbm}
\usepackage{bm}
\usepackage{fontawesome}
\usepackage{upgreek}
\usepackage{xspace}
\usepackage{cite}
\usepackage{subcaption}
\usepackage{textcomp}

\usepackage{tikz}
\usetikzlibrary{automata, positioning, arrows, matrix}

\usepackage{annotates}

\newcommand{\labeledarrow}[1]{\raisebox{-3pt}{$\xrightarrow{#1}$}}

\newcommand{\arc}[3]{\left\langle#1\labeledarrow{\text{#2}}#3\right\rangle}

\newcommand{\tran}{\leftrightarrow}
\newcommand{\emit}{\text{\scalebox{0.75}{$\nearrow$}}}
\newcommand{\pat}{\text{\scalebox{0.75}{\faSearch}}}

\newcommand{\longname}{regular-pattern-sensitive CRF}
\newcommand{\Longname}{Regular-pattern-sensitive CRF}
\newcommand{\shortname}{RPCRF\xspace}

\newcommand{\longnames}{{\longname}s\xspace}
\newcommand{\Longnames}{{\Longname}s\xspace}
\newcommand{\shortnames}{{\shortname}s\xspace}

\title{\Longnames for Distant Label Interactions}

\author{Sean Papay* \and Roman Klinger* \and Sebastian Pad\'o\dag \vspace{.5em}\\
  *Fundamentals of Natural Language Procccessing \\
  University of Bamberg, Germany\\
  \texttt{(sean.papay|roman.klinger)@uni-bamberg.de}\vspace{.5em}\\
  \dag Institute for Natural Language Processing \\
  University of Stuttgart, Germany\\
  \texttt{sebastian.pado@ims.uni-stuttgart.de}
}
  
\begin{document}
\maketitle              

\begin{abstract}
  While LLMs have grown popular in sequence labeling,
  linear-chain conditional random fields (CRFs)
  remain a popular alternative with the ability to directly model interactions between labels.
  However, the Markov assumption limits them to 
  interactions between adjacent labels.
 Weighted finite-state transducers (FSTs), in contrast, can 
model distant label--label interactions, but exact label inference is intractable in general. 
In this work, we present \longnames (\shortnames), a method of enriching standard linear-chain CRFs with the ability to learn long-distance label interactions 
through user-specified patterns.
This approach allows users to write regular-expression label patterns concisely specifying which types of interactions the model should take into account, allowing the model to learn from data whether and in which contexts these patterns occur.
The result can be interpreted alternatively as a CRF augmented with additional, non-local potentials, or as a finite-state transducer whose structure is defined by a set of easily-interpretable patterns.
Critically, 
exact training and inference are tractable for many pattern sets.
We detail how an \shortname can be automatically constructed from a set of user-specified patterns, and demonstrate the model's effectiveness on
a sequence of three synthetic sequence modeling datasets.
\end{abstract}

\section{Introduction}

Sequence labeling is a common paradigm which has provided a useful
frame to modeling many tasks in machine learning, ranging from Natural
Language Processing (e.g., part-of-speech (POS) tagging
\citep{schmid-1994-part,chiche2022part}) to protein structure prediction
\citep{wang2016protein,mukanov2022learning} and weather pattern
prediction \citep{https://doi.org/10.1029/2008WR007487}.

Sequence labeling is fundamentally a structured prediction task --
individual labels are not in general independent from one another, but
should form a coherent label sequence.  E.g., in weather pattern
prediction, while the weather at a specific time point may be
uncertain, it should still be highly correlated to the weather at
nearby time points. In part-of-speech tagging, where an individual
word like ``duck'' may have ambiguous POS in isolation, models strive
to tag all words so that they obtain a grammatical global POS
sequence.

In recent years, research in NLP, but also beyond, has been dominated
by the impressive developments in the area of neural networks.  With
the widespread success of LLM encoders such as BERT
\citep{devlin-etal-2019-bert}, a common approach is to represent the
entire input sequence in the joint latent space of such an LLM
encoder, and to make independent predictions for each token
conditioned on this joint latent representation.%
\footnote{Concretely, this would correspond to e.g.\ feeding the input
  into BERT, and using a position-wise softmax output layer.}  With a
sufficiently powerful encoder, models can try to sidestep the issue of
modeling interactions between output labels by modeling the
interactions at the level of the input sequence.

However, the success of LLMs is predicated on both practical and
conceptual factors.
\begin{itemize}
\item First, at the practical level, LLMs appear to be a class of
  learning methods that capitalize very well on the specific
  properties of natural language -- that is, the fact that most (hard)
  constraints are local, that sequences are fairly predictable, and
  that symbols are mildly ambiguous. In contrast, research has found
  that LLM-based models are not such clear success stories when
  applied to languages with different, properties, notably 'crisper'
  ones such as logics \citep{liu-etal-2024-proficient} and programming languages
  \citep{10.5555/3698900.3698947}
\item Second, LLMs work best when large amounts of data are available
  for pre-training, which again is not the case for all domains.
\item Third, there are conceptual limits according to which even
  strong encoder-based approaches to sequence modeling often cannot be certain about a prediction.
This may be due to underlying ambiguity (e.g.\ no model can be certain about the POS tags in an ambiguous sentence like ``I saw her duck.''),
limits imposed by data availability or model complexity, or simply the difficulty of the underlying task.
In such cases,  while models won't be able to always guess the correct label sequence,
they stand to benefit from explicitly modeling interactions between
labels, such that they can exclude unlikely label sequences.
\end{itemize}
For these reasons, we believe that structured prediction, with its
ability to cope with a larger typology of input languages, still
warrants investigation as a general approach to modeling interactions
between labels.

In this paper, we extend linear-chain conditional random fields (CRFs)
\citep{lafferty2001}, maybe the most established approach to modeling
label--label interactions.  Within this framework, interactions
between adjacent labels are directly modeled, but distant labels are
assumed to only interact by proxy of their intervening labels.  This
conditional independence assumption makes CRFs well-suited for
modeling local interactions between labels, but leads to difficulties
when long-distance interactions are important, such as in quotation
detection \citep{scheible2016model} but fundamentally unable to
account for more global constraints in the interest of computational
efficiency.

A related class of models are (neural) weighted finite-state transducers or FSTs \citep{mohri-1997-finite,eisner-2002-parameter,rastogi-etal-2016-weighting}.
Like CRFs, weighted FSTs define a distribution over label sequences conditioned on an input sequence, but they do so by modeling transitions through latent \textit{states}.
FSTs also obey a Markov assumption, but in their case, this is a conditional independence assumption on states, not on labels.
While the state at a given time step depends directly only on the states of neighboring time steps,
the output label at that time step may not be conditionally independent from distant output labels,
depending on the structure and weights of the underlying automaton, and which paths through that automaton might explain those labels.

This ability to model distant interactions makes weighted FSTs
more powerful than CRFs but also computationally more
demanding.
When the
underlying automaton is nondeterministic, inferring the most probable
label sequence is NP-hard \citep{10.1007/978-3-540-45257-7_2}.
Furthermore,
it is often not obvious how to chose the crucial automaton structure in order
to be sensitive to specific types of label--label interactions.

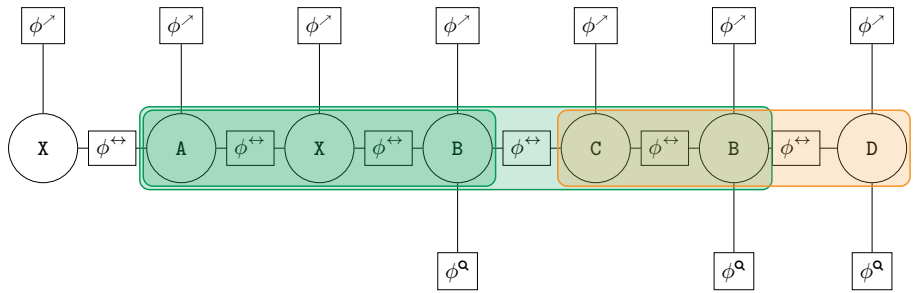
\begin{figure}[t]
  \begin{center}
    \resizebox{\columnwidth}{!}{
\begin{tikzpicture}
\tikzstyle{var}=[draw, circle, minimum size=1cm, inner sep=0cm]
\tikzstyle{given}=[fill=lightgray]
\tikzstyle{factor}=[draw, rectangle, fill=white]

\tikzstyle{match}=[rounded corners, fill opacity=0.2,line width=0.6pt]

\node[var] at (0, 0) (y1) {\texttt{X}};
\node[var] at (2, 0) (y2) {\texttt{A}};
\node[var] at (4, 0) (y3) {\texttt{X}};
\node[var] at (6, 0) (y4) {\texttt{B}};
\node[var] at (8, 0) (y5) {\texttt{C}};
\node[var] at (10, 0) (y6) {\texttt{B}};
\node[var] at (12, 0) (y7) {\texttt{D}};


\node[draw, above=of y1] (emit1) {$\phi^{\emit}$};
\node[draw, above=of y2] (emit2) {$\phi^{\emit}$};
\node[draw, above=of y3] (emit3) {$\phi^{\emit}$};
\node[draw, above=of y4] (emit4) {$\phi^{\emit}$};
\node[draw, above=of y5] (emit5) {$\phi^{\emit}$};
\node[draw, above=of y6] (emit6) {$\phi^{\emit}$};
\node[draw, above=of y7] (emit7) {$\phi^{\emit}$};

\node[draw, below=of y4] (pat4) {$\phi^{\pat}$};
\node[draw, below=of y6] (pat6) {$\phi^{\pat}$};
\node[draw, below=of y7] (pat7) {$\phi^{\pat}$};

\draw (y1) -- (emit1);
\draw (y2) -- (emit2);
\draw (y3) -- (emit3);
\draw (y4) -- (emit4);
\draw (y5) -- (emit5);
\draw (y6) -- (emit6);
\draw (y7) -- (emit7);

\draw (y4) -- (pat4);
\draw (y6) -- (pat6);
\draw (y7) -- (pat7);

\draw (y1) -- (y2) node [pos=0.5, factor] {$\phi^{\tran}$};
\draw (y2) -- (y3) node [pos=0.5, factor] {$\phi^{\tran}$};
\draw (y3) -- (y4) node [pos=0.5, factor] {$\phi^{\tran}$};
\draw (y4) -- (y5) node [pos=0.5, factor] {$\phi^{\tran}$};
\draw (y5) -- (y6) node [pos=0.5, factor] {$\phi^{\tran}$};
\draw (y6) -- (y7) node [pos=0.4, factor] {$\phi^{\tran}$};

\draw[match,fill=ForestGreen,draw=ForestGreen] (1.45, -0.55) rectangle (6.55, 0.55) {};
\draw[match,fill=ForestGreen,draw=ForestGreen] (1.4, -0.6) rectangle (10.55, 0.6) {};
\draw[match,fill=BurntOrange,draw=BurntOrange] (7.45, -0.55) rectangle (12.55, 0.55) {};

\end{tikzpicture}
}
  \end{center}
\caption{
\label{fig:intro}
A linear-chain CRF can only model probabilities of labels occurring at particular positions ($\phi^{\emit}$),
and probabilities for labels being adjacent to one another ($\phi^{\tran}$).
In particular, linear-chain CRFs cannot encourage or discourage the presence of nonlocal patterns in the label sequence, e.g.\
the regular expression patterns \textcolor{ForestGreen}{$\texttt{A}.^*\texttt{B}$} and \textcolor{BurntOrange}{$\texttt{C}.^*\texttt{D}$}.
With an \shortname, a set of such patterns can be specified, and the model can learn the probability of each of those patterns occurring at different positions of the label sequence ($\phi^{\pat}$).
}
\end{figure}

In this paper, we propose \longnames (\shortnames), a model
architecture combining the strengths of CRFs and FSTs for sequence
labeling.
An \shortname can be seen as a linear-chain CRF equipped with
the ability to be sensitive to specific types of long-distance
interactions between labels.
When instantiating a model, a user
specifies a set of regular-expression label patterns, such that the
resulting model will be able to punish or reward occurrences of those
patterns at specific positions in the label sequence. 
In this way, particular types of
long-distance interactions can be chosen in a task-specific manner,
while the model is still free to learn how and when those interactions
are important for sequence labeling.
Figure \ref{fig:intro} illustrates how an \shortname can model long-distance
interactions through sensitivity to patterns. Equivalently, \shortnames are a framework for specifying
automaton structures for FSTs in an easily interpretable manner such
that the resulting FST will be sensitive to exactly  those
long-distance interactions the user would like to model.
Unlike in
the general-case for weighted FSTs, an \shortname will always define a
deterministic automaton, support efficient exact inference like CRFs.

We first characterize \shortnames formally, and
discuss how one can be implemented as a linear-chain CRF defined over an alternative label sequence.
We then discuss the time-complexity of parameter estimation and inference.
Finally, we perform a number of experiments on synthetic data wherein we compare an \shortname against
a linear-chain, demonstrating different types of nonlocal label structures an \shortname can
be made sensitive to through an appropriate choice of patterns.

\section{Model architecture and construction}
\subsection{Formal description}
For a label set $\Upsigma$, a standard linear-chain CRF, parameterized by $\theta$, defines a distribution over
label sequences $\bm{y} \in \Upsigma^*$ conditioned  on input sequences $\bm{x}$ in terms of a
\textit{transition potential function} $\phi^{\tran}_{\theta}$ and a \textit{emission potential function} $\phi^{\emit}_{\theta}$:

\begin{equation}
P_{\theta}(\bm{y}\mid\bm{x}) = \frac{1}{Z} \prod_{i}\left(\phi^{\tran}_{\theta}(y_i, y_{i+1}) \cdot \phi^{\emit}_{\theta}(\bm{x}, y_i, i) \right)
\end{equation}
$Z$, the partition function, acts as a constant of proportionality, and is chosen such that all probabilities sum to unity:
\begin{equation}
Z = \sum_{\bm{y'}} \left(\prod_{i}\left(\phi^{\tran}_{\theta}(y'_i, y'_{i+1}) \cdot \phi^{\emit}_{\theta}(\bm{x}, y'_i, i) \right)\right)
\end{equation}
The transition potential function is applied pairwise to each pair of adjacent labels, and is responsible for modeling label-to-label interactions, while
the emission potential function models the interaction between the input sequence and individual labels.

An \shortname can be understood as standard linear-chain augmented with additional potential functions defined by the set of specified  patterns.
An \shortname is additionally hyperparameterized by a set $\mathbb{L}$ of
regular-language patterns, and includes a \textit{pattern potential
  function}, $\phi^{\pat}_{\theta}$, to model the likelihood of
different label-sequence patterns ending at different positions in the
sequence:
\begin{align}
 &  P^{\mathbb{L}}_{\theta}(\bm{y}\mid\bm{x}) \propto  P_{\theta}(\bm{y}\mid\bm{x}) \cdot \prod_{L \in \mathbb{L}} \prod_{i} \phi^{\pat}_{\theta}(L, i) ^{\mathcal I} \\
  & \text{ with } \mathcal{I} = \mathbbm{1}(L \text{ matches } \bm x \text{ ending at position } i) \nonumber
\end{align}
In principle, since deciding if an arbitrary regular-language pattern matches ending on a given label index requires looking at all preceding labels, this
defines a CRF without linear-chain structure wherein all labels are adjacent to one another. 
However, as we will show next, the \shortname distribution can be represented as the distribution over an auxiliary CRF which \textit{does} have a linear-chain structure,
allowing for tractable training and exact inference for these models.

\subsection{Construction from patterns}
This subsection describes how training and inference can be done with \shortnames.
As described, these models are highly cyclic CRFs, for which exact training and inference are infeasible in general.
However, we will present a method for defining an auxiliary, linear-chain CRF whose distribution happens to equal the \shortname distribution.
As this auxiliary CRF has a linear-chain structure, parameter estimation and inference can be done with the forward and Viterbi algorithms respectively.
 

We begin by defining a deterministic finite-state automaton (DFA) $\Uppi$ whose state space captures information about all patterns in $\mathbb{L}$.
Specifically, we would like to define $\Uppi$ such that, as $\Uppi$ processes the label sequence $\bm{y}$, the current state of $\Uppi$ at time step $i$ can tell us which set of patterns in $\mathbb{L}$ match $\bm{y}$ ending at position $i$.
We achieve this as follows: for each $L \in \mathbb{L}$, we construct a DFA for the language $L' = \Upsigma^*\oplus L$, i.e., the language of
label sequences with a suffix matching $L$.
We can then construct $\Uppi$ as a product of the automata for these $L'$, whose states are $|\mathbb{L}|$-tuples of the states the constituent automata.
While accepting $\bm{y}$ through $\Uppi$, we can examine the state-tuple at each time-step, and determine which set of patterns match $\bm{y}$ ending at that time step by checking which states in that tuple are accepting states in their original automata.
We can interpret $\Uppi$ as a state-labeled DFA, where each state is labeled with the set of patterns
which match $\bm{y}$ ending at that time-step when that state is reached.
In particular, for each state $q$ in $\Uppi$, we will notate the set of patterns which label that state as $\mathbb{L}_{[q]} \subseteq \mathbb{L}$.

\begin{figure*}[tb!]
\begin{subfigure}[b]{\textwidth}
  \begin{center}    
  \begin{tikzpicture}[->,>=stealth',shorten >=1pt,auto,node distance=2.8cm,
                    semithick]
  
  \tikzstyle{inpath}=[draw=ForestGreen, ultra thick]

  \node[initial,state] at (0, 0) (q1) {\shortstack{$q_1$\\$\varnothing$}};
  \node[state] at (3.5, 1.5) (q2)  {\shortstack{$q_2$\\$\varnothing$}};
  \node[state] at (7, 1.5) (q3)  {\shortstack{$q_3$\\\scalebox{0.7}{$\{L_1\}$}}};
  \node[state] at (3.5, -1.5) (q4)  {\shortstack{$q_4$\\$\varnothing$}};
  \node[state] at (7, -1.5) (q5)  {\shortstack{$q_5$\\\scalebox{0.7}{$\{L_2\}$}}};

  
  \path (q1)  edge [loop above] node {\texttt{X}} (q1);
  \path[preaction={draw,line width=1mm}] (q1)  edge  node {\texttt{A}} (q2);
  \path[preaction={draw,line width=1mm}] (q2)  edge [inpath,above,bend left=15]  node {\textcolor{ForestGreen}{\texttt{A}}} (q3);
  \path[preaction={draw,line width=1mm}] (q3)  edge [below,bend left=15]  node {\texttt{X}} (q2);
  \path[preaction={draw,line width=1mm}] (q2)  edge [inpath,loop above] node {\textcolor{ForestGreen}{\texttt{X}}} (q2);
  \path[preaction={draw,line width=1mm}] (q3)  edge [inpath,loop above]  node {\textcolor{ForestGreen}{\texttt{A}}} (q3);
  
  \path[preaction={draw,line width=1mm}] (q1)  edge [inpath,below]  node {\textcolor{ForestGreen}{\texttt{B}}} (q4);
  \path[preaction={draw,line width=1mm}] (q4)  edge [below,bend right=15]  node {\texttt{B}} (q5);
  \path[preaction={draw,line width=1mm}] (q5)  edge [above,bend right=15]  node {\texttt{X}} (q4);
  \path[preaction={draw,line width=1mm}] (q4)  edge [loop below] node {\texttt{X}} (q4);
  \path[preaction={draw,line width=1mm}] (q5)  edge [loop below]  node {\texttt{B}} (q5);
  
  \path[preaction={draw,line width=1mm}] (q2)  edge [bend right]  node {\texttt{B}} (q4);
  \path[preaction={draw,line width=1mm}] (q4)  edge [inpath,bend right]  node {\textcolor{ForestGreen}{\texttt{A}}} (q2);
  
  \path[preaction={draw,line width=1mm}] (q3)  edge [pos=0.2,below,bend right=0]  node {\texttt{B}} (q4);
  \path[preaction={draw,line width=1mm}] (q5)  edge [pos=0.2,above,bend left=0] node {\texttt{A}} (q2);


\end{tikzpicture}
  \end{center}
\caption{\label{fig:dfa}
A DFA for for $\Uppi$. The path through this automaton for the string \texttt{BAXAA} is marked.}
\end{subfigure}
\begin{subfigure}[b]{\textwidth}

\begin{multline*}
A = \{
\arc{q_1}{\texttt{X}}{q_1}, \arc{q_1}{\texttt{A}}{q_2}, \arc{q_1}{\texttt{B}}{q_4}, \arc{q_2}{\texttt{X}}{q_2}, \arc{q_2}{\texttt{A}}{q_3}, \arc{q_2}{\texttt{B}}{q_4},
\arc{q_3}{\texttt{X}}{q_2}, \\ \arc{q_3}{\texttt{A}}{q_3}, \arc{q_3}{\texttt{B}}{q_4}, \arc{q_4}{\texttt{A}}{q_2}, \arc{q_4}{\texttt{X}}{q_4}, \arc{q_4}{\texttt{B}}{q_5},
\arc{q_4}{\texttt{A}}{q_2}, \arc{q_4}{\texttt{X}}{q_3}, \arc{q_4}{\texttt{B}}{q_4}
\}
\end{multline*}
\caption{$A$, the set of arcs in $\Uppi$, which will be used as the label set for the auxiliary CRF.}
\end{subfigure}
\begin{subfigure}[b]{\textwidth}
  \begin{center}    
\resizebox{0.75\textwidth}{!}{
  \begin{tikzpicture}
\tikzstyle{var}=[draw, circle, minimum size=1cm, inner sep=0cm]
\tikzstyle{given}=[fill=lightgray]
\tikzstyle{factor}=[draw, rectangle, fill=white]

\node[var] at (0, 0) (y1) {$\arc{q_1}{\texttt{B}}{q_4}$};
\node[var] at (3.25, 0) (y2) {$\arc{q_4}{\texttt{A}}{q_2}$};
\node[var] at (6.5, 0) (y3) {$\arc{q_2}{\texttt{X}}{q_2}$};
\node[var] at (9.75, 0) (y4) {$\arc{q_2}{\texttt{A}}{q_3}$};
\node[var] at (13, 0) (y5) {$\arc{q_3}{\texttt{A}}{q_3}$};
\node[var, given] at (6.5, 4) (x) {$\bm{x}$};
\draw (x) -- (y1) node [pos=0.5, factor] {$\phi^{\emit}_{\theta}(\bm{x},\text{B}, 1)$};
\draw (x) -- (y2) node [pos=0.8, factor] {$\phi^{\emit}_{\theta}(\bm{x},\text{A}, 2)$};
\draw (x) -- (y3) node [pos=0.5, factor] {$\phi^{\emit}_{\theta}(\bm{x},\text{X}, 3)$};
\draw (x) -- (y4) node [pos=0.8, factor] {$\phi^{\emit}_{\theta}(\bm{x},\text{A}, 4) \cdot \phi^{\pat}_\theta(L_1, 4)$};
\draw (x) -- (y5) node [right,pos=0.5, factor] {$\phi^{\emit}_{\theta}(\bm{x},\text{A}, 5) \cdot \phi^{\pat}_\theta(L_1, 5)$};
\draw (y1) -- (y2) node [midway, factor] {$\phi^{\tran}_{\theta}(\text{B}, \text{A})$};
\draw (y2) -- (y3) node [midway, factor] {$\phi^{\tran}_{\theta}(\text{A}, \text{X})$};
\draw (y3) -- (y4) node [midway, factor] {$\phi^{\tran}_{\theta}(\text{X}, \text{A})$};
\draw (y4) -- (y5) node [midway, factor] {$\phi^{\tran}_{\theta}(\text{A}, \text{A})$};
\end{tikzpicture}
}
\end{center}
\caption{The auxiliary CRF calculating the probability for the arc sequence corresponding to $\bm{y}$'s path through $\Uppi$.
Since $q_3$ corresponds to an accepting state for $L_1$, the emission function incorporates the pattern potential for $L_1$ at time steps which end on $q_3$. 
The resulting probability equals the \shortname probability for the string $\bm{y}$.}
\end{subfigure}
\caption{%
\label{fig:eg}
A worked example for the label string $\bm{y} = \texttt{BAXAA}$ of an \shortname with two patterns: $L_1 = \texttt{A}\texttt{X}^*\texttt{A}$ and $L_2 = \texttt{B}\texttt{X}^*\texttt{B}$.
(a) shows $\Uppi$, the state-labeled automaton we obtain from these two languages, (b) shows the set of arcs in $\Uppi$, which will be tags for our auxiliary CRF, and (c) demonstrates how we use our auxiliary CRF to calculate a probability for $\bm{y}$.
}
\end{figure*}
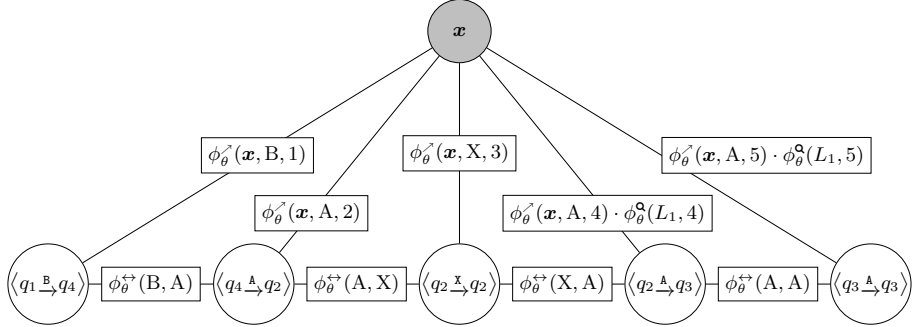

Once we have constructed $\Uppi$, we will define an auxiliary linear-chain CRF whose label set is the set $A$ of arcs (labeled arrows) of $\Uppi$.
As $\Uppi$ is deterministic, each possible label sequence $\bm{y} \in \Upsigma^*$ corresponds to exactly one path through $\Uppi$ -- as a path through $\Uppi$ can
be represented as a sequence of arcs $\bm{\pi} \in A^*$ , that path can be used directly as a label sequence for our auxiliary CRF.
We specifically construct our auxiliary CRF such that the probability assigned to each arc sequence $\bm{\pi}$ is equal to the
\shortname probability for the corresponding label sequence $\bm{y}$:
\begin{align}
  P'_{\theta}(\bm{\pi} \mid \bm{x})& =  \frac{1}{Z} \prod_{i}\left(\phi'^{\tran}_{\theta}(\pi_i, \pi_{i+1}) \cdot \phi'^{\emit}_{\theta}(\bm{x}, \pi_i, i) \right) \nonumber \\
  & = P^{\mathbb{L}}_{\theta}(\bm{y} \mid \bm{x})
\end{align}
We achieve this through suitable definition of our auxiliary CRF's transition function $\phi'^{\tran}_{\theta}$ and emission function $\phi'^{\emit}_{\theta}$:
\begin{equation}
\phi'^{\tran}_{\theta}(\arc{q}{a}{r},\arc{s}{b}{t}) = \begin{cases}
\phi^{\tran}_{\theta}(a, b) & \text{if } r = s \\
0 & \text{otherwise}\\
\end{cases}
\end{equation}

\begin{align}
  \phi'^{\emit}_{\theta}(\bm{x}, \arc{q}{a}{r}, i) = 
  \begin{cases}
  0 & \text{if } \mathcal{C} \\
  \phi^{\emit}_{\theta}(\bm{x}, a, i) \cdot \\
  \prod\limits_{L \in \mathbb{L}_{[r]}} \phi^{\pat}_{\theta}(L, i) & \text{otherwise}
\end{cases} \\
  \text{where }\mathcal{C} = \mathbbm{1}(i = 1 \text{ and $q$ is not initial state of $\Uppi$}) \nonumber
\end{align}
These definitions ensure that our auxiliary CRF will only assign nonzero probability to proper paths through $\Uppi$ (which start at the initial
state and contain only valid transitions), and, for those paths, will assign a probability to path $\bm{\pi}$ equal to the \shortname distribution's probability for the
corresponding label sequence $\bm{y}$.
Figure~\ref{fig:eg} shows a worked example of this construction, illustrating the state-labeled automaton
obtained from a set of patterns and the auxiliary CRF computing a probability for a path through that automaton.

As the time- and space-complexity of our learning and inference
algorithms will depend on the size of $\Uppi$, we would like to make
$\Uppi$ as small as possible.
This can be achieved by minimizing all automata for our $L'$ languages before constructing $\Uppi$,
and pruning unreachable states in $\Uppi$.%

In the worst case, all states in $\Uppi$ will be reachable, and the size of $\Uppi$ equals the product
of the minimal number of states for all languages in $\mathbb{L}$, i.e.\ it is exponential in $|\mathbb{L}|$.
However, we observe that in many cases where different patterns ``share information,'' we can do significantly better than this upper bound.
For instance, when one pattern is a strict prefix of another, we can include the prefix pattern ``for free'', without necessitating any additional states,
as the product construction has the effect  of simply labeling which states in the larger automaton match the prefix.
Unfortunately, a full characterization of such synergies falls outside the scope of the current work.

\section{Experiments}
To concretely demonstrate the differences between \shortnames and linear-chain CRFs, we perform three experiments with synthetic data,
each demonstrating a particular class of problem where an \shortname can model interactions not capturable by a linear-chain CRF.
Each experiment will feature a synthetic dataset exhibiting a certain type of label structure, and a pattern set designed to be sensitive to that label structure.
As all labels are trivially independent under certainty (i.e. when all label probabilities are either zero or one), all synthetic data tasks are fundamentally
underspecified, such that models will always need to ``guess'' the right answer from some space of possibilities.
Thus, for each experiment, in addition to reporting model performance, we will report the highest level of performance possible by a hypothetical model employing an optimal strategy.

For all synthetic data experiments, we will use digits as input symbols, and letters and underscores as output labels, with the specific
meanings of these symbols varying by experiment.
For all experiments, the emission and pattern potential functions are represented with a biLSTM neural network \citep{hochreiter1997long},
and the transition function is represented as a parameter matrix.
All parameters are jointly optimized until convergence using the Adam optimizer \citep{adam}.

We evaluate all tasks via exact-match accuracy.
That means that we count a model as correct only when it predicts the label sequence exactly correct, and we don't assign partial credit.
This turns out to be quite important, as many less-strict evaluation methods are explicitly insensitive to the global structures we are trying to capture.
For instance, when evaluating by token-wise accuracy, models are not rewarded for producing globally plausible label sequences, only for ensuring that each individual label is likely in isolation, something that linear-chain CRFs are already capable of.

\subsection{Experiment 1: Cardinality patterns}
A common source of label interdependencies in sequence labeling is given by global constraints on how often a particular label occurs.
Under such constraints, each label can directly depend on each other
label.  For example, if we know that a particular label must occur
exactly once in a sequence, assigning that label to any particular
position affects the marginal distribution of every other position.
These constraints may be soft, though -- for example, in the
classification of daily activities from a smartwatch data sequence,
users typically go running once a day, but might run twice, or not at
all \citep{kwon18}.


In order to investigate an \shortname's ability to model such cardinality constraints,
we construct a synthetic dataset of $(\bm{x}, \bm{y})$ pairs.
For each pair, $\bm{x}$ consists of a single non-zero digit $k$, followed by nine zeros.
The first label of $\bm{y}$ is always \texttt{\_}, and, of the remaining nine labels, exactly $k$ are \texttt{A}, with all others being \texttt{\_}.
We chose the value of $k$ uniformly randomly, and then uniformly randomly select which $k$ positions should be labeled as \texttt{A}.

\begin{table}[t]
\center
\caption{Example for Experiment 1 (cardinality patterns). The first token of each input specifies the number of \texttt{A}s in the output.}
\label{tab:cardeg}
\resizebox{\columnwidth}{!}{
  \begin{tabular}{lrrr}
\toprule
$\bm{x}$ & \texttt{  3000000000  } & \texttt{9000000000  } & \texttt{1000000000} \\
$\bm{y}$ & \texttt{  \_\_A\_AA\_\_\_\_  } & \texttt{\_AAAAAAAAA  } & \texttt{\_\_\_\_\_A\_\_\_\_}\\
\bottomrule
\end{tabular}}
\end{table}

As patterns, we use a set of nine regular languages $\mathbb{L} = \{L_1, \cdots, L_9\}$:
\begin{equation}
L_k = \texttt{\^{}}(\texttt{\_}^*\texttt{a})^k\texttt{\_}^*\texttt{\$}
\end{equation}
Each $L_k$ matches label sequences with exactly $k$ occurrences of \texttt{A}.
As pattern can match only a complete label sequence, and as the languages are disjoint, only one pattern can match any given label sequence.
An \shortname should be able to learn from the first token of the input sequence which pattern should apply to the label sequence,
and assign only that pattern a high weight with its pattern potential function,
resulting in the model always predicting the correct number of \texttt{A}s.
Conversely, while a CRF can learn that the \texttt{A} label should be more or less
likely depending on the value of $k$, it has no mechanism for enforcing a specific number of $\texttt{A}$ labels
(except in the case for $k=9$, wherein the output is deterministic).


\begin{table}[t]
\center
\caption{\label{tab:rescard} 
  Results for Experiment~1 (EM acc. = Exact-match accuracy; Opt. str. = optimal strategy).
}
\begin{tabular}{lrr}
\toprule
Model & EM acc. (\%) & \% Opt. str.\\
\midrule
Optimal strategy& 14.64 & --\\
\midrule
LSTM+CRF        & 11.27 & 76.98 \\
LSTM+\shortname & 14.61 & 99.80 \\
\bottomrule
\end{tabular}
\end{table}
Table~\ref{tab:cardeg} gives examples of some datapoints for this experiment.
Table~\ref{tab:rescard} summarizes the performance of \shortname and linear-chain CRFs on this task.
We see that an \shortname is able to achieve near-optimal accuracy.
On the other hand, the linear-chain CRF, unable to directly enforce cardinality constraints, can only achieve approximately 77\% of the optimal strategy's accuracy.

\subsection{Experiment 2: Agreement patterns}
Commonly for sequence labeling tasks, the presence of one type of label in a sequence might be highly informative
about the presence or absence of other labels at distant positions in the sequence.
For instance, when using sequence labeling to label named entities in text, 
an entity of type \textsc{Event} may be likely to occur in the same document as an entity of type \textsc{Date},
while there may be no such affinity  between entities of types \textsc{Law} and \textsc{Work\_of\_art}.
In the extreme case, certain labels might be guaranteed to co-occur in a document, or alternatively
forbidden from doing so.

To investigate an \shortname's ability to learn such interactions, we construct a synthetic sequence-labeling
dataset which exhibits strong agreement interactions between distant labels.
In each $(\bm{x}, \bm{y})$ pair, $\bm{x}$ is a length-ten sequence containing eight zeros and exactly two ones, which represent entities to be labeled.
The corresponding $\bm{y}$ assigns a \texttt{\_} label to all zeros, and a letter from $\texttt{A}$ to $\texttt{F}$ to the two ones.
Importantly, these letter labels are selected such that $\texttt{A}$ must co-occur with a $\texttt{B}$, $\texttt{C}$ with a $\texttt{D}$,
and $\texttt{E}$ with an $\texttt{F}$. Table \ref{tab:exp2-example} provides some example $(\bm{x}, \bm{y})$-pairs for this experiment.

\begin{table}[t]
\center
\caption{\label{tab:exp2-example}
Example for Experiment~2 (agreement patterns): model must learn which pairs of non-zero output labels correspond (A/B, C/D, E/F).}
\resizebox{\columnwidth}{!}{
\begin{tabular}{lrrr}
\toprule
$\bm{x}$ & \texttt{  0010000100  } & \texttt{0011000000  } & \texttt{0001000001} \\
$\bm{y}$ & \texttt{  \_\_A\_\_\_\_B\_\_  } & \texttt{\_\_DC\_\_\_\_\_\_  } & \texttt{\_\_\_F\_\_\_\_\_E}\\
\bottomrule
\end{tabular}}
\end{table}

We assume a setting where model users know that \textit{some} co-occurrence constraints exist, but do not know the particular
letters which can or cannot co-occur.
Thus, as patterns, we use a set of ${6 \choose 2} = 15$ languages, with each language matching a label sequence containing two distinct labels exactly once:
\begin{align}
\mathbb{L} = \Big\{& \texttt{\^{}}\texttt{\_}^*\left(\alpha\texttt{\_}^*\beta \mid \beta\texttt{\_}^*\alpha\right)\texttt{\_}^*\texttt{\$}
                     : \nonumber \\
  &\left\{\alpha, \beta\right\} \subseteq \left\{ \texttt{A}, \texttt{B}, \texttt{C}, \texttt{D}, \texttt{E}, \texttt{F} \right\}, \alpha \neq \beta\Big\}
\end{align}
Our model is thus responsible for learning which label pairs agree and disagree with one another.

Table~\ref{tab:exp2-results} shows the results on this experiment for an \shortname and for a linear-chain CRF baseline.
As before, our \shortname-based model achieves nearly optimal performance, while the linear-chain CRF, unable to learn the relationships between distant labels, lags significantly behind.
Interestingly, the linear-chain CRF is able to model agreement in \textit{some} cases -- namely when the two entities happen to be directly adjacent Due to this, it performs better than the $\frac{1}{36}$ odds we would expect from having it label the two entities independently, but fails in cases where the
entities are distant from one another.

\begin{table}[t]
\center
\caption{\label{tab:exp2-results}
  Results for Experiment 2 on agreement patterns
  (EM acc. = Exact-match accuracy; Opt. str. = optimal strategy).
}
\begin{tabular}{lrr}
\toprule
Model & EM acc. (\%) & \% Opt. str.\\
\midrule
Optimal strategy & 16.67 & --\\
\midrule
LSTM+CRF         &  6.97 & 41.81\\
LSTM+\shortname  & 16.60 &  99.58\\

\bottomrule
\end{tabular}
\end{table}

\subsection{Experiment 3: Battleship}
While this paper has thus-far focused largely on CRFs with a linear-chain structure, CRFs are also commonly used for 2-dimensional data in tasks such as image segmentation \citep{chen2017deeplab}.
In such a setting, instead of labeling elements of a sequence, individual pixels or grid cells are labeled.
Crucially, such a setting usually envisions each pixel as directly adjacent to all four of its orthogonal neighbors, leading to a highly cyclic graph structure not amenable to tractable exact inference \citep{loopy}.

\begin{table*}[th!]
\center
\caption{\label{tab:egship}
Example for Experiment 3 (battleship). Each input marks a single
cell of the battleship, while the output marks all of its cells. Inputs/outputs are shown as $5\times5$ grids here but 
are treated as length-25 sequences by models.}
\small
\begin{tabular}{l@{\hskip 1cm}r@{\hskip 1cm}r@{\hskip 1cm}r}
\toprule
$\bm{x}$ &
	\begin{tabular}{lllll}
		\texttt{0}&\texttt{0}&\texttt{0}&\texttt{0}&\texttt{0}\\
		\texttt{0}&\texttt{0}&\texttt{0}&\texttt{0}&\texttt{0}\\
		\texttt{0}&\texttt{0}&\texttt{0}&\texttt{1}&\texttt{0}\\
		\texttt{0}&\texttt{0}&\texttt{0}&\texttt{0}&\texttt{0}\\
		\texttt{0}&\texttt{0}&\texttt{0}&\texttt{0}&\texttt{0}\\
	\end{tabular} &
	\begin{tabular}{lllll}
		\texttt{0}&\texttt{0}&\texttt{0}&\texttt{0}&\texttt{0}\\
		\texttt{0}&\texttt{0}&\texttt{0}&\texttt{0}&\texttt{0}\\
		\texttt{1}&\texttt{0}&\texttt{0}&\texttt{0}&\texttt{0}\\
		\texttt{0}&\texttt{0}&\texttt{0}&\texttt{0}&\texttt{0}\\
		\texttt{0}&\texttt{0}&\texttt{0}&\texttt{0}&\texttt{0}\\
	\end{tabular} &
	\begin{tabular}{lllll}
		\texttt{0}&\texttt{0}&\texttt{0}&\texttt{0}&\texttt{0}\\
		\texttt{0}&\texttt{0}&\texttt{0}&\texttt{0}&\texttt{0}\\
		\texttt{1}&\texttt{0}&\texttt{0}&\texttt{0}&\texttt{0}\\
		\texttt{0}&\texttt{0}&\texttt{0}&\texttt{0}&\texttt{0}\\
		\texttt{0}&\texttt{0}&\texttt{0}&\texttt{0}&\texttt{0}\\
	\end{tabular} \\
	\midrule
$\bm{y}$ &
	\begin{tabular}{lllll}
		\texttt{\_}&\texttt{\_}&\texttt{\_}&\texttt{A}&\texttt{\_}\\
		\texttt{\_}&\texttt{\_}&\texttt{\_}&\texttt{A}&\texttt{\_}\\
		\texttt{\_}&\texttt{\_}&\texttt{\_}&\texttt{A}&\texttt{\_}\\
		\texttt{\_}&\texttt{\_}&\texttt{\_}&\texttt{A}&\texttt{\_}\\
		\texttt{\_}&\texttt{\_}&\texttt{\_}&\texttt{\_}&\texttt{\_}\\
	\end{tabular} &
	\begin{tabular}{lllll}
		\texttt{\_}&\texttt{\_}&\texttt{\_}&\texttt{\_}&\texttt{\_}\\
		\texttt{\_}&\texttt{\_}&\texttt{\_}&\texttt{\_}&\texttt{\_}\\
		\texttt{A}&\texttt{A}&\texttt{A}&\texttt{A}&\texttt{\_}\\
		\texttt{\_}&\texttt{\_}&\texttt{\_}&\texttt{\_}&\texttt{\_}\\
		\texttt{\_}&\texttt{\_}&\texttt{\_}&\texttt{\_}&\texttt{\_}\\
	\end{tabular}&
	\begin{tabular}{lllll}
		\texttt{\_}&\texttt{\_}&\texttt{\_}&\texttt{\_}&\texttt{\_}\\
		\texttt{A}&\texttt{\_}&\texttt{\_}&\texttt{\_}&\texttt{\_}\\
		\texttt{A}&\texttt{\_}&\texttt{\_}&\texttt{\_}&\texttt{\_}\\
		\texttt{A}&\texttt{\_}&\texttt{\_}&\texttt{\_}&\texttt{\_}\\
		\texttt{A}&\texttt{\_}&\texttt{\_}&\texttt{\_}&\texttt{\_}\\
	\end{tabular}\\
\bottomrule
\end{tabular}
\end{table*}

With appropriate encoding and patterns, \shortnames can also be used for labeling such 2-dimensional data.
Any 2-dimensional grid can be serialized row-by-row into a linear sequence.
Cells which neighbored horizontally in the original grid are still neighbors in the sequence, while vertical neighbors are now separated
by from one another by a constant distance equal to the grid width.
By writing patterns that are specifically sensitive to labels separated by exactly this distance, we can
enable an \shortname to model interactions between vertically adjacent cells in our original grid.

We demonstrate this concretely with a synthetic task on a $5\times5$ grid.
Somewhere on this grid, a $4\times1$ battleship is hiding, positioned and oriented randomly.
The input sequence $\bm{x}$ comprises all zeros, except for a single one, at some randomly-chosen cell of the battleship.
In the label sequence $\bm{y}$, each cell occupied by the battleship is labeled \texttt{A}, while all other cells are labeled \texttt{\_}.
The model's task is thus to guess the position and location of the battleship, given only a single ``hit.''

Table \ref{tab:egship} illustrates some input-output pairs.
We use a single pattern, sensitive to two \texttt{A}s separated by
four \texttt{\_}s (i.e., vertically adjacent in the grid):
\begin{equation}
\mathbb{L} = \{\texttt{A\_\_\_\_A}\}
\end{equation}
This allows \shortname to be sensitive to vertically adjacent pairs of \texttt{A}s in the label sequence (at least when all intervening labels are instances of \texttt{\_}).

\begin{table}[t]
\center
\caption{\label{tab:exp3-results}  
  Results for Experiment 3, Battleship
    (EM acc. = Exact-match accuracy; Opt. str. = optimal strategy).
}
\begin{tabular}{lrr}
\toprule
Model & EM acc. (\%) & \% Opt. str.\\
\midrule
Optimal strategy & 31.25 & --\\
\midrule
LSTM+CRF         &  2.50  & 8.00 \\
LSTM+\shortname  & 12.49 & 39.98 \\
\bottomrule
\end{tabular}
\end{table}

Table~\ref{tab:exp3-results} reports the performance of our two models. In this case, the \shortname-based model does not
achieve the performance of the optimal strategy here.  This is due to
a limitation in the pattern used: while the model can use its pattern
to ensure the predicted $\bm{A}$s are adjacent, it has no mechanism
for ensuring that it predicts the correct \textit{number} of
$\bm{A}$s.  Nonetheless, even though the provided pattern set cannot
capture all structural properties of the label sequences, we still see
significant improvements over a linear-chain CRF.


\section{Related Work}
Our proposed approach is one of many ways for extending a linear-chain
CRF in a manner that selectively circumvents the Markov assumption of
default CRFs.  Here we will briefly discuss some alternate formalisms
for defining and working with such 'higher-order' CRFs.

\paragraph{Pattern-based CRFs.}
A conceptually similar approach to our current proposal are pattern-based CRFs
\citep{NIPS2009_94f6d7e0,takhanov2013inference}.
As with our \longnames, pattern-based CRFs allow practitioners to specify a set of label patterns,
allowing the CRF to learn long-distance dependencies by either encouraging or discouraging the presence of these patterns
at particular locations of the label sequence.
However, the patterns in pattern-based CRFs are limited to exact string matches, while our \shortnames allow for arbitrary regular-expression patterns.
Critically, a pattern-based CRF can only model dependencies as distant as its longest search pattern, while \shortnames can easily be designed to learn dependencies over arbitrary distances, as our Experiment~1 demonstrated.

\paragraph{Semi-Markov CRFs.}
Another approach commonly used for allowing CRFs to learn non-local label interactions are semi-Markov CRFs \citep{sarawagi2004semi}.
Under this formalism, rather than labeling each individual token, a semi-Markov CRF outputs a segmentation of the input, labeling each segment.
While segment labels must follow the Markov assumption (each segment's label depends directly only on its neighboring segments), the
model's behavior \textit{within} each segment may be non-Markovian.
Such models offer an efficient approach to modeling certain types of nonlocal interactions, but these interactions are limited to occurring within the same segment, again in contrast to our model.

\paragraph{Skip-chain CRFs.}
A skip-chain CRFs \citep{sutton2007introduction} is an otherwise
linear-chain CRF augmented with \textit{skip-connections}, a number of
connections directly connecting otherwise distant labels in the
sequence. The exact structure of these skip connections can be
specified according to the task, and may even be specified conditioned
on the input sequence. This provides  a conceptually straightforward way to enable linear-chain CRFs to model long-distance dependencies.
While skip connections can be selected to account for many possible types of long-distance interactions, the resulting graphs are highly cyclic, and often require approximate techniques for parameter estimation and inference.
Nonetheless, with certain connection structures, tricks are possible to allow for exact training and inference on skip-chain CRFs \citep{galley2006skip}.  

\paragraph{Regular-constrained CRFs.}
Regular-con\-strain\-ed CRFs \citep{papay2021constraining} enforce that a
model's output sequence \textit{must} match some
user-specified regular expression.
While this enables linear-chain
CRFs to respect non-local label interactions,
our proposal allows a CRF to learn the likelihood of regular
expressions matching at different positions in the label sequence.
Thus, a regular-constrained CRF can be understood as a special case of
a \shortname with a single pattern (the complement of the
user-specified language) given a constant potential of zero.
While regular-constrained CRFs are limited to enforcing constraints
known a priori, our \longnames can \textit{learn} when different label
patterns are likely or unlikely.

\section{Conclusions}

This paper introduced \longnames, a method for enriching linear-chain
CRFs with the ability to learn long-distance interactions which occur
within user-specified regular-expression patterns.  By representing
all patterns in a single state-labeled DFA, and using an auxiliary CRF
to represent a distribution over paths through this DFA, we can
selectively extend CRFs with non-local
features while preserving efficient parameter learning and inference.

Regular patterns are often sufficient to model the relevant structures
in the domain, as Experiment~2 illustrates. More complex structures
can often be rewritten with regular patterns by assuming a maximum
input length (cf. \citep{mohri2001regular} and Experiment~1).
Even when regular-language patterns cannot fully capture the dependency
structure of the labels, and imperfect approximation can still yield a substantial improvement,
as we found in Experiment~3.

Regular patterns offer a flexible and powerful tool for incorporating
domain knowledge into sequence classification models that combine the
knowledge-based and data-driven paradigms in a promising fashion.
Sequence labeling models can be made to
account for specific tasks' output structures by simply specifying 
regular-expression patterns, without the need to explicitly
construct an FST or otherwise adapt the model architecture.

A promising direction for future work lies in the combination of \shortnames
with LLM encoders.
The strengths of these two paradigms could prove complementary, and
LLMs with \shortname output layers may make good models for structured prediction tasks
such as relation extraction or semantic role labeling, where it is necessary
to model both linguistic interactions in the input as well as structural interactions
in the output.


  


%
%
%

\section*{Limitations}
While training and inference time for \shortnames are quadratic in the number
of arcs in the underlying automaton, this number is worst-case exponential in the number of patterns, limiting
our model's use with some large sets of patterns.
While some combinations of patterns synergize and yield small automata, we do not have a formal characterization of
which combinations of patterns lead to tractable models.

\bibliography{lit}

\begin{thebibliography}{26}
\providecommand{\natexlab}[1]{#1}

\bibitem[{Casacuberta and de~la Higuera(2000)}]{10.1007/978-3-540-45257-7_2}
Francisco Casacuberta and Colin de~la Higuera. 2000.
\newblock Computational complexity of problems on probabilistic grammars and
  transducers.
\newblock In \emph{Grammatical Inference: Algorithms and Applications}, pages
  15--24, Berlin, Heidelberg. Springer Berlin Heidelberg.

\bibitem[{Chen et~al.(2017)Chen, Papandreou, Kokkinos, Murphy, and
  Yuille}]{chen2017deeplab}
Liang-Chieh Chen, George Papandreou, Iasonas Kokkinos, Kevin Murphy, and Alan~L
  Yuille. 2017.
\newblock Deeplab: Semantic image segmentation with deep convolutional nets,
  atrous convolution, and fully connected crfs.
\newblock \emph{IEEE transactions on pattern analysis and machine
  intelligence}, 40(4):834--848.

\bibitem[{Chiche and Yitagesu(2022)}]{chiche2022part}
Alebachew Chiche and Betselot Yitagesu. 2022.
\newblock Part of speech tagging: a systematic review of deep learning and
  machine learning approaches.
\newblock \emph{Journal of Big Data}, 9(1):10.

\bibitem[{Devlin et~al.(2019)Devlin, Chang, Lee, and
  Toutanova}]{devlin-etal-2019-bert}
Jacob Devlin, Ming-Wei Chang, Kenton Lee, and Kristina Toutanova. 2019.
\newblock {BERT}: Pre-training of deep bidirectional transformers for language
  understanding.
\newblock In \emph{Proceedings of the 2019 Conference of the North {A}merican
  Chapter of the Association for Computational Linguistics: Human Language
  Technologies, Volume 1 (Long and Short Papers)}, pages 4171--4186,
  Minneapolis, Minnesota. Association for Computational Linguistics.

\bibitem[{Eisner(2002)}]{eisner-2002-parameter}
Jason Eisner. 2002.
\newblock \href {https://doi.org/10.3115/1073083.1073085} {Parameter estimation
  for probabilistic finite-state transducers}.
\newblock In \emph{Proceedings of the 40th Annual Meeting of the Association
  for Computational Linguistics}, pages 1--8, Philadelphia, Pennsylvania, USA.
  Association for Computational Linguistics.

\bibitem[{Fang et~al.(2024)Fang, Miao, Srivastav, Liu, Zhang, Fang, Asmita,
  Tsang, Nazari, Wang, and Homayoun}]{10.5555/3698900.3698947}
Chongzhou Fang, Ning Miao, Shaurya Srivastav, Jialin Liu, Ruoyu Zhang, Ruijie
  Fang, Asmita, Ryan Tsang, Najmeh Nazari, Han Wang, and Houman Homayoun. 2024.
\newblock Large language models for code analysis: do llms really do their job?
\newblock In \emph{Proceedings of the 33rd USENIX Conference on Security
  Symposium}, SEC '24, USA. USENIX Association.

\bibitem[{Galley(2006)}]{galley2006skip}
Michel Galley. 2006.
\newblock A skip-chain conditional random field for ranking meeting utterances
  by importance.
\newblock In \emph{Proceedings of the 2006 Conference on Empirical Methods in
  Natural Language Processing}, pages 364--372, Sydney, Australia. Association
  for Computational Linguistics.

\bibitem[{Hochreiter and Schmidhuber(1997)}]{hochreiter1997long}
Sepp Hochreiter and J{\"u}rgen Schmidhuber. 1997.
\newblock Long short-term memory.
\newblock \emph{Neural computation}, 9(8):1735--1780.

\bibitem[{Kingma and Ba(2015)}]{adam}
Diederik~P. Kingma and Jimmy Ba. 2015.
\newblock Adam: {A} method for stochastic optimization.
\newblock In \emph{Proceedings of the International Conference on Learning
  Representations}, San Diego, CA.

\bibitem[{Kwon and Choi(2018)}]{kwon18}
Min-Cheol Kwon and Sunwoong Choi. 2018.
\newblock \href {https://doi.org/10.1155/2018/2618045} {Recognition of daily
  human activity using an artificial neural network and smartwatch}.
\newblock \emph{Wireless Communications and Mobile Computing}.

\bibitem[{Lafferty et~al.(2001)Lafferty, McCallum, and Pereira}]{lafferty2001}
John~D. Lafferty, Andrew McCallum, and Fernando C.~N. Pereira. 2001.
\newblock Conditional {Random} {Fields:} {Probabilistic} {Models} for
  {Segmenting} and {Labeling} {Sequence} {Data}.
\newblock In \emph{Proceedings of the Eighteenth International Conference on
  Machine Learning}, pages 282--289.

\bibitem[{Liu et~al.(2024)Liu, Cao, Shi, Zhang, Nie, Hu, Hou, and
  Li}]{liu-etal-2024-proficient}
Jinxin Liu, Shulin Cao, Jiaxin Shi, Tingjian Zhang, Lunyiu Nie, Linmei Hu, Lei
  Hou, and Juanzi Li. 2024.
\newblock \href {https://doi.org/10.18653/v1/2024.findings-acl.45} {How
  proficient are large language models in formal languages? an in-depth insight
  for knowledge base question answering}.
\newblock In \emph{Findings of the Association for Computational Linguistics:
  ACL 2024}, pages 792--815, Bangkok, Thailand. Association for Computational
  Linguistics.

\bibitem[{Mohri(1997)}]{mohri-1997-finite}
Mehryar Mohri. 1997.
\newblock Finite-state transducers in language and speech processing.
\newblock \emph{Computational Linguistics}, 23(2):269--311.

\bibitem[{Mohri and Nederhof(2001)}]{mohri2001regular}
Mehryar Mohri and Mark-Jan Nederhof. 2001.
\newblock Regular approximation of context-free grammars through
  transformation.
\newblock In \emph{Robustness in language and speech technology}, pages
  153--163. Springer.

\bibitem[{Mukanov and Takhanov(2022)}]{mukanov2022learning}
Zhalgas Mukanov and Rustem Takhanov. 2022.
\newblock Learning the pattern-based {CRF} for prediction of a protein local
  structure.
\newblock \emph{Informatica}, 46(6).

\bibitem[{Murphy et~al.(1999)Murphy, Weiss, and Jordan}]{loopy}
Kevin~P. Murphy, Yair Weiss, and Michael~I. Jordan. 1999.
\newblock Loopy belief propagation for approximate inference: An empirical
  study.
\newblock In \emph{Proceedings of the Fifteenth Conference on Uncertainty in
  Artificial Intelligence}, page 467–475, Stockholm, Sweden. Morgan Kaufmann
  Publishers Inc.

\bibitem[{Papay et~al.(2022)Papay, Klinger, and Pado}]{papay2021constraining}
Sean Papay, Roman Klinger, and Sebastian Pado. 2022.
\newblock Constraining linear-chain {CRF}s to regular languages.
\newblock In \emph{International Conference on Learning Representations}.

\bibitem[{Raje and Mujumdar(2009)}]{https://doi.org/10.1029/2008WR007487}
Deepashree Raje and P.~P. Mujumdar. 2009.
\newblock \href
  {https://arxiv.org/abs/https://agupubs.onlinelibrary.wiley.com/doi/pdf/10.1029/2008WR007487}
  {A conditional random field–based downscaling method for assessment of
  climate change impact on multisite daily precipitation in the mahanadi
  basin}.
\newblock \emph{Water Resources Research}, 45(10).

\bibitem[{Rastogi et~al.(2016)Rastogi, Cotterell, and
  Eisner}]{rastogi-etal-2016-weighting}
Pushpendre Rastogi, Ryan Cotterell, and Jason Eisner. 2016.
\newblock \href {https://doi.org/10.18653/v1/N16-1076} {Weighting finite-state
  transductions with neural context}.
\newblock In \emph{Proceedings of the 2016 Conference of the North {A}merican
  Chapter of the Association for Computational Linguistics: Human Language
  Technologies}, pages 623--633, San Diego, California. Association for
  Computational Linguistics.

\bibitem[{Sarawagi and Cohen(2004)}]{sarawagi2004semi}
Sunita Sarawagi and William~W Cohen. 2004.
\newblock Semi-markov conditional random fields for information extraction.
\newblock In \emph{Advances in Neural Information Processing Systems},
  volume~17. MIT Press.

\bibitem[{Scheible et~al.(2016)Scheible, Klinger, and
  Pad{\'o}}]{scheible2016model}
Christian Scheible, Roman Klinger, and Sebastian Pad{\'o}. 2016.
\newblock \href {https://doi.org/10.18653/v1/P16-1164} {Model architectures for
  quotation detection}.
\newblock In \emph{Proceedings of the 54th Annual Meeting of the Association
  for Computational Linguistics (Volume 1: Long Papers)}, pages 1736--1745,
  Berlin, Germany. Association for Computational Linguistics.

\bibitem[{Schmid(1994)}]{schmid-1994-part}
Helmut Schmid. 1994.
\newblock Part-of-speech tagging with neural networks.
\newblock In \emph{Proceedings of the 15th {I}nternational {C}onference on
  {C}omputational {L}inguistics}, Kyoto, Japan.

\bibitem[{Sutton and McCallum(2007)}]{sutton2007introduction}
Charles Sutton and Andrew McCallum. 2007.
\newblock An introduction to conditional random fields for relational learning.

\bibitem[{Takhanov and Kolmogorov(2013)}]{takhanov2013inference}
Rustem Takhanov and Vladimir Kolmogorov. 2013.
\newblock Inference algorithms for pattern-based crfs on sequence data.
\newblock In \emph{Proceedings of the International Conference on Machine
  Learning}, pages 145--153.

\bibitem[{Wang et~al.(2016)Wang, Peng, Ma, and Xu}]{wang2016protein}
Sheng Wang, Jian Peng, Jianzhu Ma, and Jinbo Xu. 2016.
\newblock Protein secondary structure prediction using deep convolutional
  neural fields.
\newblock \emph{Scientific reports}, 6(1):1--11.

\bibitem[{Ye et~al.(2009)Ye, Lee, Chieu, and Wu}]{NIPS2009_94f6d7e0}
Nan Ye, Wee Lee, Hai Chieu, and Dan Wu. 2009.
\newblock Conditional random fields with high-order features for sequence
  labeling.
\newblock In \emph{Proceedings of Advances in Neural Information Processing
  Systems}. Curran Associates, Inc.

\end{thebibliography}

\end{document}